\documentclass[runningheads]{llncs}
\usepackage{graphicx}
\usepackage{balance}
\usepackage{amsmath}
\usepackage[tight,footnotesize]{subfigure}
\usepackage{bm}
\usepackage{algorithm}
\usepackage[noend]{algorithmic}
\usepackage{multirow}
\usepackage{amsfonts}

\begin{document}
\title{The Power of Second Chance: Personalized Submodular Maximization with Two Candidates}
\titlerunning{Personalized Submodular Maximization with Two Candidates}
\author{Jing Yuan\inst{1} \and Shaojie Tang\inst{2}}
\institute{
Department of Computer Science and Engineering, University of North Texas \and Department of Management Science and Systems, School of Management, University at Buffalo}

\maketitle              % typeset the header of the contribution
\begin{abstract}
Most of existing studies on submodular maximization focus on selecting a subset of items that maximizes a \emph{single} submodular function. However, in many real-world scenarios, we might have multiple user-specific functions, each of which models the utility of a particular type of user. In these settings, our goal would be to choose a set of items that performs well across all the user-specific functions. One way to tackle this problem is to select a single subset that maximizes the sum of all of the user-specific functions. Although this aggregate approach is efficient in the sense that it avoids computation of sets for individual functions, it really misses the power of personalization - for it does not allow to choose different sets for different functions. In this paper, we introduce the problem of personalized submodular maximization with two candidate solutions. For any two candidate solutions, the utility of each user-specific function is defined as the better of these two candidates. Our objective is, therefore, to select the best set of two candidates that maximize the sum of utilities of all the user-specific functions. We have designed effective algorithms for this problem. We also discuss how our approach generalizes to multiple candidate solutions, increasing flexibility and personalization in our solution.
\end{abstract}
\section{Introduction}
A submodular function is defined by its intuitive diminishing returns property: adding an item to a smaller set will increase the return more in comparison with when this happens from a larger set. Such a function is extremely common in various combinatorial optimization problems naturally arising from machine learning, graph theory, economics, and game theory.  Most of the work in submodular optimization focuses on selecting a subset of items from a ground set  that maximizes a single submodular function. However, in many real-world scenarios, we are confronted with multiple user-specific functions denoted as $f_1, \cdots, f_m : 2^\Omega \rightarrow \mathbb{R}_{\geq 0}$. Each of these functions, such as $f_i$, captures the utility corresponding to some user type indexed by $i$. Our main goal will be to maximize the aggregate utility of all the $m$ functions. One trivial way to achieve this would be to compute a solution individually for every single function $f_i$. Unfortunately, this would require to compute and store $m$ solutions, which is infeasible or at least very inefficient if the number of user-specific functions is large.
Another way is to look for a \emph{single} feasible solution, denoted as $S\subseteq \Omega$, that maximizes the summation of these $m$ functions, i.e., $\max_{S\subseteq \Omega}\sum_{i\in[m]} f_i(S)$. This problem, also known as the maximization of decomposable submodular functions \cite{schwartzman2024mini}, has been well-studied in the literature and efficient algorithms have been designed for the same. Nevertheless, such an aggregate approach, despite being efficient, is unable to harness the power of personalization.  Specifically, it does not provide the flexibility in offering a personalized set for each function.

In our research, we introduce the innovative concept of personalized submodular maximization. Consider a pair of sets $\{S_1, S_2\}$, for each user-specific function $f_i$, we determine its utility based on the better-performing solution among these two candidates, represented as $\max\{f_i(S_1), f_i(S_2)\}$. Mathematically, our problem can be expressed as follows:
\begin{eqnarray*}
&&\max_{S_1, S_2 \subseteq \Omega}\sum_{i\in[m]} \max\{f_i(S_1), f_i(S_2)\} \\
&&\mbox{ subject to } |S_1|\leq k, |S_2|\leq k,
\end{eqnarray*}
where $k$ is the size constraint of a feasible solution.
In essence, our primary objective is to maximize the combined utility of user-specific functions while maintaining a personalized approach to item selection. An important and practical application of our study is in the context of two-stage optimization. Here, we consider that $f_1, \cdots, f_m$ represent training examples of functions drawn from an unknown distribution, we aim to choose a pair of candidate solutions based on these $m$ functions, ensuring that  one of the chosen candidates performs well when faced with a new function from the same distribution.

In this paper, we also discuss the possibility of expanding our approach  to accommodate multiple (more than two) candidate solutions. This potential extension would further enhance the flexibility and personalization options within our solution.

\subsection{Related Work} The problem of submodular maximization has received considerable attention in the literature \cite{gharan2011submodular,buchbinder2014submodular,tang2021beyond,tang2021pointwise,tang2022group}. For example, one of the most well-established results is that a simple greedy algorithm achieves a tight approximation ratio of $(1-1/e)$ for maximizing a single monotone submodular function subject to cardinality constraints \cite{nemhauser1978analysis}. Since most datasets are so big nowadays, several works were devoted to reducing the running time to maximize a submodular function. Examples include the development of accelerated greedy algorithms  \cite{mirzasoleiman2015lazier} and streaming algorithms \cite{badanidiyuru2014streaming}. All of these works, however, focus on finding a single set that maximizes a submodular function. In contrast, our goal is to identify a pair of candidates that maximizes the sum of the better-performing solution between them. This presents a unique challenge, as the resulting objective function is no longer submodular. Consequently, existing results on submodular optimization cannot be directly applied to our study.

Our work is closely related to the field of two-stage submodular optimization \cite{balkanski2016learning,mitrovic2018data,stan2017probabilistic,tang2023data}, in which the key objective is to find a smaller ground set from a large one. This reduction should be designed in such a way that choosing the items from the small set guarantees approximately the same performance as choosing items from the original large set for a variety of submodular functions. This aligns with our objective of seeking two initial solutions that cut down on computational effort in optimization with a new function. However, problem formulations between our studies are largely different despite sharing the same objective. Thereby, new methodologies should be developed to cope with the distinctive challenges presented in our research. Moreover, note that in the traditional framework of two-stage submodular optimization, once a reduced ground set is computed, further optimization based on this reduced set usually involves algorithms with possibly high time complexity, such as the greedy algorithm. In contrast, our personalized optimization model requires only a comparison between the performance of two candidate solutions, significantly reducing the computational burden in the second stage.
\section{Problem Formulation}
Our problem involves an input set of $n$ items denoted as $\Omega$, and a collection of $m$ submodular functions, namely, $f_1, \cdots, f_m : 2^\Omega \rightarrow \mathbb{R}_{\geq 0}$. To clarify, the notation $\Delta_i(x, A)$ denotes the marginal gain of adding item $x$ to set $A$ with respect to the function $f_i$. That is, $\Delta_i(x, A)=f_i(\{x\}\cup A) - f_i(A)$. Specifically, a function $f_i$ is considered submodular if and only if $\Delta_i(x, A) \geq \Delta_i(x, A')$ holds for any two sets $A$ and $A'$ where $A\subseteq A'\subseteq \Omega$ and for any item $x\in \Omega$ such that $x\notin A'$.

Our aim is to select a pair of candidate solutions, $S_1$ and $S_2$, and the utility of each user-specific function is determined by the superior solution among these two candidates.  These subsets should provide good performance across all $m$ functions when we are limited to choosing solutions from either $S_1$ or $S_2$. Formally,

 \begin{center}
\framebox[0.7\textwidth][c]{
\enspace
\begin{minipage}[t]{0.7\textwidth}
\small
$\textbf{P.0}$
$\max_{S_1, S_2 \subseteq \Omega}\sum_{i\in[m]} \max\{f_i(S_1), f_i(S_2)\} $\\
$\mbox{ subject to } |S_1|\leq k, |S_2|\leq k$,
\end{minipage}
}
\end{center}
\vspace{0.1in}
where $k$ is the size constraint of a feasible solution.

A straightforward approach to solving $\textbf{P.0}$ is to transform it into a standard set selection problem. Specifically, we can introduce a ground set $\mathcal{U} = \{(i,j)\mid i\in \Omega, j\in \{1, 2\}\}$. Here, selecting an element $(i,j)\in \mathcal{U}$ corresponds to placing item $i$ in set $S_j$ in our original problem. Let $x_{ij}$ be a binary decision variable representing the selection of $(i,j)$, such that $x_{ij}=1$ if and only if $(i,j)$ is selected. Then $\textbf{P.0}$ is reduced to finding a set of elements from $\mathcal{U}$ such that $\forall i\in\Omega, x_{i1}+x_{i2}=1$ and $\forall j\in\{1, 2\}, \sum_{i\in \Omega}x_{ij} \leq k$, which represents the intersection of two matroid constraints. Unfortunately, it is straightforward to verify that the utility function defined over $\mathcal{U}$ is not necessarily submodular, even if each individual function $f_i$ is submodular. Hence, existing solutions for submodular maximization subject to two matroid constraints are not directly applicable to our problem.

%In general, $\max\{f_i(S_1), f_i(S_2)\}$ and, consequently, $\sum_{i\in[m]} \max\{f_i(S_1), f_i(S_2)\}$, do not form submodular functions, even if each individual $f_i$ is submodular. As a result, the existing findings and techniques in submodular optimization may not be directly applicable to our specific problem $\textbf{P.0}$.  In the rest of this paper, let $F(S_1, S_2)=\sum_{i\in[m]} \max\{f_i(S_1), f_i(S_2)\}$ denote the objective function of $\textbf{P.0}$.

\section{Algorithm Design for Constant $m$}
\label{sec:small}
We first study the case if the number of functions $m$ is a constant. Before presenting our algorithm, we introduce a new optimization problem $\textbf{P.1}$. The objective of this problem is to partition the $m$ functions  into two groups such that the sum of the optimal solutions for these two groups is maximized. Formally,

 \begin{center}
\framebox[0.7\textwidth][c]{
\enspace
\begin{minipage}[t]{0.7\textwidth}
\small
$\textbf{P.1}$
\begin{eqnarray*}\max_{A, B\subseteq [m]} &&\big(\max_{S\subseteq \Omega: |S|\leq k} \sum_{i\in A} f_i(S) + \max_{S\subseteq \Omega: |S|\leq k} \sum_{i\in B} f_i(S)\big)
\end{eqnarray*}
subject to $B = [m]\setminus A$.
\end{minipage}
}
\end{center}
\vspace{0.1in}

We next show that the optimal solution of $\textbf{P.1}$ serves as an upper bound for our original problem.
\begin{lemma}
\label{lem:67}
Let $OPT_{1}$ (resp. $OPT_{0}$) denote the value of the optimal solution of $\textbf{P.1}$ (resp. our original problem $\textbf{P.0}$), we have
\begin{eqnarray}
OPT_{1} \geq OPT_{0}.
\end{eqnarray}
\end{lemma}
\emph{Proof:} Assume $S_1^*$ and $S_2^*$ is the optimal solution of $\textbf{P.0}$, we can partition $m$ functions to two groups $A'$ and $B'$ such that every function in $A'$ favors $S_1^*$ and  every function in $B'$ favors $S_2^*$. That is,
\[A'=\{i\in [m]\mid f_i(S_1^*) \geq f_i(S_2^*)\}\] and \[B'=\{i\in [m]\mid f_i(S_1^*) < f_i(S_2^*)\}.\] Hence,
\begin{eqnarray*}
&&OPT_{0}= \sum_{i\in[m]} \max\{f_i(S_1^*), f_i(S_2^*)\} \\
&&= \sum_{i\in A'}  \max\{f_i(S_1^*), f_i(S_2^*)\} + \sum_{i\in B'}  \max\{f_i(S_1^*), f_i(S_2^*)\}\\
&&= \sum_{i\in A'}  f_i(S_1^*) + \sum_{i\in B'}  f_i(S_2^*)
\end{eqnarray*}
where the first inequality is by the definition of $OPT_{0}$, the second equality is by the observation that $A'$ and $B'$ is a partition of $[m]$ and the third equality is by the definitions of $A'$ and $B'$.

Moreover, it is easy to verify that
\begin{eqnarray*}
&&\max_{S\subseteq \Omega: |S|\leq k} \sum_{i\in A'} f_i(S) + \max_{S\subseteq \Omega: |S|\leq k} \sum_{i\in B'} f_i(S) \\
&&\geq  \sum_{i\in A'} f_i(S_1^*) + \sum_{i\in B'} f_i(S_2^*).
\end{eqnarray*} This is because $|S_1^*|\leq k$ and $|S_2^*|\leq k$. It follows that
\begin{eqnarray*}
&&OPT_{0}= \sum_{i\in A'}  f_i(S_1^*) + \sum_{i\in B'}  f_i(S_2^*) \\
&&\leq \max_{S\subseteq \Omega: |S|\leq k} \sum_{i\in A'} f_i(S) + \max_{S\subseteq \Omega: |S|\leq k} \sum_{i\in B'} f_i(S).
\end{eqnarray*}

Therefore,
\begin{eqnarray*}
&&OPT_{1} = \\
&&\max_{A, B\subseteq [m]} \big(\max_{S\subseteq \Omega: |S|\leq k} \sum_{i\in A} f_i(S) + \max_{S\subseteq \Omega: |S|\leq k} \sum_{i\in B} f_i(S)\big) \\
&&\geq \max_{S\subseteq \Omega: |S|\leq k} \sum_{i\in A'} f_i(S) + \max_{S\subseteq \Omega: |S|\leq k} \sum_{i\in B'} f_i(S)\\
&&\geq \sum_{i\in A'}  f_i(S_1^*) + \sum_{i\in B'}  f_i(S_2^*) = OPT_{0}.
\end{eqnarray*} This finishes the proof of this lemma. $\Box$

Now, we present our algorithm, called \textsf{Enumeration-based Algorithm}, which is listed in Algorithm \ref{alg:LPP0}. Our approach involves enumerating all possible partitions of $[m]$. For each partition, denoted as $A$ and $B$, we utilize a state-of-the-art algorithm to solve two subproblems: \[\max_{S\subseteq \Omega: |S|\leq k} \sum_{i\in A} f_i(S)\] and \[\max_{S\subseteq \Omega: |S|\leq k} \sum_{i\in B} f_i(S).\] This results in obtaining two sets, $C_1$ and $C_2$, respectively. Finally, we return the best pair of sets as the solution for our original problem $\textbf{P.0}$.

Since the number of functions $m$ is a constant, the maximum number of possible partitions we must enumerate is at most $O(2^m)$, which is also a constant. As long as $\max_{S\subseteq \Omega: |S|\leq k} \sum_{i\in A} f_i(S)$ and $\max_{S\subseteq \Omega: |S|\leq k} \sum_{i\in B} f_i(S)$ can be solved in polynomial time, the \textsf{Enumeration-based Algorithm} is a polynomial time algorithm.  Next we provide an approximation ratio of  Algorithm \ref{alg:LPP0}.

\begin{algorithm}[hptb]
\caption{\textsf{Enumeration-based Algorithm}}
\label{alg:LPP0}
\begin{algorithmic}[1]
\STATE $S_1\leftarrow \emptyset, S_2\leftarrow \emptyset$
\FOR { $A \subseteq [m]$}
\STATE $B\leftarrow [m]\setminus A$
\STATE  $C_1 \leftarrow$ $\alpha$-approximation solution of \[\max_{S\subseteq \Omega: |S|\leq k} \sum_{i\in A} f_i(S)\] %\label{line:1}
\STATE $C_2 \leftarrow$ $\alpha$-approximation solution of \[\max_{S\subseteq \Omega: |S|\leq k} \sum_{i\in B} f_i(S)\] %\label{line:2}
\IF{ \[\sum_{i\in[m]} \max\{f_i(C_1), f_i(C_2)\} \geq \sum_{i\in[m]} \max\{f_i(S_1), f_i(S_2)\}\]}
\STATE $(S_1, S_2)\leftarrow (C_1, C_2)$
\ENDIF
\ENDFOR
\RETURN $S_1, S_2$
\end{algorithmic}
\end{algorithm}

\begin{lemma}
\label{lem:c}
Assuming the existence of $\alpha$-approximation algorithms for \[\max_{S\subseteq \Omega: |S|\leq k} \sum_{i\in A} f_i(S)\] for any $A \subseteq [m]$, our \textsf{Enumeration-based Algorithm} (Algorithm \ref{alg:LPP0}) provides an $\alpha$-approximation solution for $\textbf{P.0}$.
\end{lemma}
\emph{Proof:} Assuming that $A^*$ and $B^*$ represent the optimal solution for $\textbf{P.1}$, let us consider the round of our algorithm where it enumerates the partition of $A^*$ and $B^*$. In this round, we denote the solutions obtained as $C_1$ and $C_2$. Given that there exist $\alpha$-approximation algorithms  for $\max_{S\subseteq \Omega: |S|\leq k} \sum_{i\in A} f_i(S)$ for any $A \subseteq [m]$, by adopting this algorithm as a subroutine, we have \[\sum_{i\in A^*} f_i(C_1) \geq \alpha \max_{S\subseteq \Omega: |S|\leq k} \sum_{i\in A^*} f_i(S)\] and \[\sum_{i\in B^*} f_i(C_2)\geq \alpha \max_{S\subseteq \Omega: |S|\leq k} \sum_{i\in B^*} f_i(S).\] Hence,
\begin{eqnarray*}
&&\sum_{i\in[m]} \max\{f_i(C_1), f_i(C_2)\}\geq \sum_{i\in A^*} f_i(C_1)  + \sum_{i\in B^*} f_i(C_2)\\
&&\geq \alpha \big(\max_{S\subseteq \Omega: |S|\leq k} \sum_{i\in A^*} f_i(S) +  \max_{S\subseteq \Omega: |S|\leq k} \sum_{i\in B^*} f_i(S)\big)\\
&&=  \alpha OPT_{1}
\end{eqnarray*}
where the equality is by the assumption that $A^*$ and $B^*$ represent the optimal solution for $\textbf{P.1}$.

This, together with Lemma \ref{lem:67}, implies that
\begin{eqnarray}
\sum_{i\in[m]} \max\{f_i(C_1), f_i(C_2)\}\geq  \alpha OPT_{1} \geq \alpha OPT_{0}.
\end{eqnarray}

This lemma is a consequence of the above inequality and the fact that the final solution obtained by our algorithm is at least as good as $\sum_{i\in[m]} \max\{f_i(C_1), f_i(C_2)\}$. $\Box$

Observe that if all $f_i$ are monotone and submodular functions, then there exists $(1-1/e)$-approximation algorithms for $\max_{S\subseteq \Omega: |S|\leq k} \sum_{i\in A} f_i(S)$ for any $A \subseteq [m]$. Therefore, by substituting $\alpha=1-1/e$ into Lemma \ref{lem:c}, we obtain the following theorem.
\begin{theorem}
\label{lem:c1}
Assume all  $f_i$ are monotone and submodular functions, \textsf{Enumeration-based Algorithm} (Algorithm \ref{alg:LPP0}) provides an $(1-1/e)$-approximation solution for $\textbf{P.0}$.
\end{theorem}

\section{Algorithm Design for Large $m$}
\label{sec:large}
When dealing with a large value of $m$, relying on an enumeration-based approach can become impractical. In this section, we introduce a \textsf{Sampling-based Algorithm}, outlined in Algorithm \ref{alg:LPP1}, that provides provable performance bounds. Instead of exhaustively enumerating all possible partitions of $[m]$, we examine $T$ \emph{random} partitions. For each partition, we follow the same procedure as in Algorithm \ref{alg:LPP0} to compute two candidate solutions. Specifically, for each sampled partition, we employ a state-of-the-art $\alpha$-approximation algorithm to solve two subproblems. Ultimately, we return the best pair of sets as the final solution.

\begin{algorithm}[hptb]
\caption{\textsf{Sampling-based Algorithm}}
\label{alg:LPP1}
\begin{algorithmic}[1]
\STATE $S_1\leftarrow \emptyset, S_2\leftarrow \emptyset, T$
\FOR { $t\in [T]$}
\STATE Randomly sample a subset of functions $A \subseteq [m]$
\STATE $B\leftarrow [m]\setminus A$
\STATE  $C_1 \leftarrow$ $\alpha$-approximation solution of \[\max_{S\subseteq \Omega: |S|\leq k} \sum_{i\in A} f_i(S)\] %\label{line:1}
\STATE $C_2 \leftarrow$ $\alpha$-approximation solution of \[\max_{S\subseteq \Omega: |S|\leq k} \sum_{i\in B} f_i(S)\] %\label{line:2}
\IF{ \[\sum_{i\in[m]} \max\{f_i(C_1), f_i(C_2)\}  \geq \sum_{i\in[m]} \max\{f_i(S_1), f_i(S_2)\} \]}
\STATE $(S_1, S_2)\leftarrow (C_1, C_2)$
\ENDIF
\ENDFOR
\RETURN $S_1, S_2$
\end{algorithmic}
\end{algorithm}

In the following two lemmas, we provide two performance bounds for Algorithm \ref{alg:LPP1}. The first bound is independent of the number of samples $T$; thus, it holds even if $T=1$. The second bound depends on $T$, increasing as $T$ increases.
\begin{lemma}
\label{lem:b}
Assuming the existence of $\alpha$-approximation algorithms for \[\max_{S\subseteq \Omega: |S|\leq k} \sum_{i\in A} f_i(S)\] for any $A \subseteq [m]$, our \textsf{Sampling-based Algorithm} (Algorithm \ref{alg:LPP1}) provides an $\alpha/2$-approximation solution for $\textbf{P.0}$.
\end{lemma}
\emph{Proof:} We first recall some notations form the proof of Lemma \ref{lem:67}. Assume $S_1^*$ and $S_2^*$ is the optimal solution of $\textbf{P.0}$, we partition all $m$ functions  to two groups $A'$ and $B'$ such that every function in $A'$ favors $S_1^*$ and  every function in $B'$ favors $S_2^*$. That is, \[A'=\{i\in [m]\mid f_i(S_1^*) \geq f_i(S_2^*)\}\] and \[B'=\{i\in [m]\mid f_i(S_1^*) < f_i(S_2^*)\}.\] Without loss of generality, we assume that $\sum_{i\in A'} f_i(S_1^*)\geq \sum_{i\in B'} f_i(S_2^*)$, implying that $\sum_{i\in A'} f_i(S_1^*)\geq OPT_{0}/2$. Now, let us consider any arbitrary partition sample denoted as $A$ and $B$, generated by our algorithm, we have
\begin{eqnarray*}
&&\max_{S\subseteq \Omega: |S|\leq k} \sum_{i\in A} f_i(S) +  \max_{S\subseteq \Omega: |S|\leq k} \sum_{i\in B} f_i(S)~\nonumber\\
&&\geq \sum_{i\in A} f_i(S_1^*) + \sum_{i\in B} f_i(S_1^*) \geq \sum_{i\in A'} f_i(S_1^*) \geq OPT_{0}/2 \label{eq:99}
\end{eqnarray*}
where the first inequality is by the observation that $|S_1^*|\leq k$, the second inequality is by the observation that $A' \subseteq A\cup B$ and the third inequality is by the observation that $\sum_{i\in A'} f_i(S_1^*)\geq OPT_{0}/2$. Because there exist $\alpha$-approximation algorithms  for $\max_{S\subseteq \Omega: |S|\leq k} \sum_{i\in A} f_i(S)$ for any $A \subseteq [m]$, by adopting this algorithm as a subroutine to compute $C_1$ and $C_2$, we have
\[\sum_{i\in A} f_i(C_1) \geq \alpha \max_{S\subseteq \Omega: |S|\leq k} \sum_{i\in A} f_i(S)\] and \[\sum_{i\in B} f_i(C_2)\geq \alpha \max_{S\subseteq \Omega: |S|\leq k} \sum_{i\in B} f_i(S).\]  Hence,
\begin{eqnarray*}
&&\sum_{i\in[m]} \max\{f_i(C_1), f_i(C_2)\} \geq \sum_{i\in A} f_i(C_1)  + \sum_{i\in B} f_i(C_2)\\
&&\geq \alpha \big(\max_{S\subseteq \Omega: |S|\leq k} \sum_{i\in A} f_i(S) +  \max_{S\subseteq \Omega: |S|\leq k} \sum_{i\in B} f_i(S)\big)\geq  (\alpha/2) OPT_{0}
\end{eqnarray*}
where the third inequality is by inequality (\ref{eq:99}). This finishes the proof of this lemma. $\Box$

\begin{lemma}
\label{lem:a}
Assuming the existence of $\alpha$-approximation algorithms for \[\max_{S\subseteq \Omega: |S|\leq k} \sum_{i\in A} f_i(S)\] for any $A \subseteq [m]$, our \textsf{Sampling-based Algorithm} (Algorithm \ref{alg:LPP1}), after $T$ rounds, provides an $\alpha \gamma(T)  (\frac{1}{2} + \frac{\epsilon}{\sqrt{m}})$-approximation solution for $\textbf{P.0}$ in expectation where $\gamma(T)=1-(\frac{1}{2}+\epsilon\frac{e}{\pi})^T$.
\end{lemma}
\emph{Proof:} Consider an arbitrary round of our algorithm, and let $A$ and $B$ denote the sampled partition, and let $(C_1, C_2)$ denote the solution returned from this round. Observe that
\begin{eqnarray*}
\sum_{i\in[m]} \max\{f_i(C_1), f_i(C_2)\} \geq \sum_{i\in A} f_i(C_1)  + \sum_{i\in B} f_i(C_2).
\end{eqnarray*}

Hence, the expected value of $\sum_{i\in[m]} \max\{f_i(C_1), f_i(C_2)\}$, where the expectation is taken over $A, B$, is at least $\mathbb{E}_{A, B}[\sum_{i\in A} f_i(C_1) + \sum_{i\in B} f_i(C_2)]$. Recall that our algorithm runs $T$ rounds and returns the best $(C_1,C_2)$ as the final solution, to prove this lemma, it suffices to show that the expected value of $\sum_{i\in[m]} \max\{f_i(C_1), f_i(C_2)\}$ is at least $\alpha \gamma(T)  (\frac{1}{2} + \frac{\epsilon}{\sqrt{m}})OPT_{0}$. To achieve this, we will focus on proving
$\mathbb{E}_{A, B}[\sum_{i\in A} f_i(C_1) + \sum_{i\in B} f_i(C_2)]\geq \alpha \gamma(T)  (\frac{1}{2} + \frac{\epsilon}{\sqrt{m}})OPT_{0}$. The rest of the proof is devoted to proving this inequality.

First,
\begin{eqnarray}
&&\mathbb{E}_{A, B}[\sum_{i\in A} f_i(C_1) + \sum_{i\in B} f_i(C_2)]~\nonumber\\
&&\geq \mathbb{E}_{A, B}[\alpha\max_{S\subseteq \Omega: |S|\leq k} \sum_{i\in A} f_i(S) +  \alpha \max_{S\subseteq \Omega: |S|\leq k} \sum_{i\in B} f_i(S)]~\nonumber\\
&&= \alpha \mathbb{E}_{A, B}[\max_{S\subseteq \Omega: |S|\leq k} \sum_{i\in A} f_i(S) +  \max_{S\subseteq \Omega: |S|\leq k} \sum_{i\in B} f_i(S)]~\nonumber\\
&&=\alpha \mathbb{E}_{A}[\max_{S\subseteq \Omega: |S|\leq k} \sum_{i\in A} f_i(S)] + \alpha \mathbb{E}_{B}[\max_{S\subseteq \Omega: |S|\leq k} \sum_{i\in B} f_i(S)]~\nonumber\\
&&\geq \alpha \mathbb{E}_{A}[ \sum_{i\in A} f_i(S_1^*)] + \alpha \mathbb{E}_{B}[\sum_{i\in B} f_i(S_2^*)]. \label{eq:sb}
\end{eqnarray}

Next, we provide lower bounds for $\mathbb{E}_{A}[ \sum_{i\in A} f_i(S_1^*)]$ and $\mathbb{E}_{B}[\sum_{i\in B} f_i(S_2^*)]$. Recall that we defined $A'=\{i\in [m]\mid f_i(S_1^*) \geq f_i(S_2^*)\}$ and $B'=\{i\in [m]\mid f_i(S_1^*) < f_i(S_2^*)\}$. Now, for some $\beta\in[0,1]$, let us denote the event as $E$, which occurs when the following condition holds for at least one partition $(A, B)$ that is enumerated by our algorithm: $\frac{|A\cap A'|}{|A'|}\geq \beta$. Because each item of $A'$ is included in $A$ independently with a probability of $1/2$, for any $\beta\in[0,1]$, we have the following:
\begin{eqnarray}
\label{eq:aa}
 \mathbb{E}_{A}[ \sum_{i\in A} f_i(S_1^*)] \geq \Pr[\mathbf{1}_E=1]\cdot \beta  \sum_{i\in A'} f_i(S_1^*).
\end{eqnarray}
%Similarly, because each item of $B'$ is included in $B$ independently with a probability of $1/2$, for any $\beta\in[0,1]$, we have the following:
%\begin{eqnarray}
%\label{eq:bb}
% \mathbb{E}_{B}[ \sum_{i\in A} f_i(S_2^*)] \geq \Pr[\frac{|B\cap B'|}{|B'|}\geq \beta]\cdot \beta  \sum_{i\in B'} f_i(S_2^*).
%\end{eqnarray}

Consider a random sample $A$ from $[m]$ and observe that each item of $A'$ is included in $A$ independently with a probability of $1/2$, by an ``anti-concentration'' result on binomial distributions (Lemma 22.2 in \cite{lecture}), we have
\begin{eqnarray*}
 \Pr[|A\cap A'| \geq \frac{|A'|}{2}+\epsilon \sqrt{|A'|}] \geq \frac{1}{2} - \epsilon\frac{e}{\pi}.
\end{eqnarray*}
%\begin{eqnarray}
% \Pr[|B\cap B'| \geq \frac{|B'|}{2}+\epsilon \sqrt{|B'|}] \geq \frac{1}{2} - \epsilon\frac{e}{\pi}.
%\end{eqnarray}
This implies that
\begin{eqnarray*}
\Pr[\frac{|A\cap A'|}{|A'|}\geq \frac{1}{2} + \frac{\epsilon}{\sqrt{|A'|}}] \geq \frac{1}{2} - \epsilon\frac{e}{\pi}.
\end{eqnarray*}
Given that $|A'| \leq m$, we further have
\begin{eqnarray*}
\Pr[\frac{|A\cap A'|}{|A'|}\geq \frac{1}{2} + \frac{\epsilon}{\sqrt{m}}] \geq \frac{1}{2} - \epsilon\frac{e}{\pi}.
\end{eqnarray*}

If we set $\beta=\frac{1}{2} + \frac{\epsilon}{\sqrt{m}}$, then we can establish a lower bound on the probability of event $E$ occurring after $T$ rounds as follows:
\begin{eqnarray*}
&&\Pr[\mathbf{1}_E=1] \geq 1-(1-\Pr[\frac{|A\cap A'|}{|A'|}\geq \frac{1}{2} + \frac{\epsilon}{\sqrt{m}}])^T \\
&&\geq 1-(\frac{1}{2}+\epsilon\frac{e}{\pi})^T.
\end{eqnarray*}
%\begin{eqnarray}
%\Pr[\frac{|B\cap B'|}{|B'|}\geq \frac{1}{2}+\frac{\epsilon}{\sqrt{|B'|}}] \geq \frac{1}{2} - \epsilon\frac{e}{\pi}.
%\end{eqnarray}

This, together with inequalities (\ref{eq:aa}), implies that
\begin{eqnarray*}
\mathbb{E}_{A}[ \sum_{i\in A} f_i(S_1^*)] \geq \Pr[\mathbf{1}_E=1]\cdot \beta  \sum_{i\in A'} f_i(S_1^*)
\geq (1-(\frac{1}{2}+\epsilon\frac{e}{\pi})^T) \cdot (\frac{1}{2} + \frac{\epsilon}{\sqrt{m}})  \sum_{i\in A'} f_i(S_1^*).
\end{eqnarray*}

Following the same argument, we can prove that
\begin{eqnarray}
\label{eq:bb1}
\mathbb{E}_{B}[ \sum_{i\in B} f_i(S_2^*)]\geq  (1-(\frac{1}{2}+\epsilon\frac{e}{\pi})^T) \cdot (\frac{1}{2} + \frac{\epsilon}{\sqrt{m}})  \sum_{i\in B'} f_i(S_2^*).
\end{eqnarray}

Let $\gamma(T)=1-(\frac{1}{2}+\epsilon\frac{e}{\pi})^T$. The above two inequalities, together with inequality (\ref{eq:sb}), imply that
\begin{eqnarray*}
&&\mathbb{E}_{A, B}[\sum_{i\in A} f_i(C_1) + \sum_{i\in B} f_i(C_2)]\geq \alpha \mathbb{E}_{A}[ \sum_{i\in A} f_i(S_1^*)] + \alpha \mathbb{E}_{B}[\sum_{i\in B} f_i(S_2^*)]\\
&&\geq \alpha \gamma(T) (\frac{1}{2} + \frac{\epsilon}{\sqrt{m}})  \sum_{i\in A'} f_i(S_1^*) + \alpha   \gamma(T) (\frac{1}{2} + \frac{\epsilon}{\sqrt{m}})  \sum_{i\in B'} f_i(S_2^*)\\
&&= \alpha \gamma(T)  (\frac{1}{2} + \frac{\epsilon}{\sqrt{m}})( \sum_{i\in A'} f_i(S_1^*)+\sum_{i\in B'} f_i(S_2^*))\\
&&=   \alpha \gamma(T)  (\frac{1}{2} + \frac{\epsilon}{\sqrt{m}}) OPT_{0}.
\end{eqnarray*} This finishes the proof of this lemma. $\Box$

By selecting a tighter bound derived from Lemma \ref{lem:b} and Lemma \ref{lem:a}, we can establish the following corollary.

\begin{corollary}
\label{cor:1}
Assuming the existence of $\alpha$-approximation algorithms for \[\max_{S\subseteq \Omega: |S|\leq k} \sum_{i\in A} f_i(S)\] for any $A \subseteq [m]$, our Sampling-based algorithm (Algorithm \ref{alg:LPP1}), after $T$ rounds, provides an $\max\{1/2,  \gamma(T)  (\frac{1}{2} + \frac{\epsilon}{\sqrt{m}})\}\cdot\alpha$-approximation solution for $\textbf{P.0}$ in expectation where $\gamma(T)=1-(\frac{1}{2}+\epsilon\frac{e}{\pi})^T$.
\end{corollary}

Observe that if all $f_i$ are monotone and submodular functions, then there exists $(1-1/e)$-approximation algorithms for $\max_{S\subseteq \Omega: |S|\leq k} \sum_{i\in A} f_i(S)$ for any $A \subseteq [m]$. Therefore, substituting $\alpha=1-1/e$ into Corollary \ref{cor:1}, we derive the following theorem.
\begin{theorem}
\label{lem:c2}
Assume all  $f_i$ are monotone and submodular functions, Sampling-based algorithm (Algorithm \ref{alg:LPP1}), after $T$ rounds, provides an $\max\{1/2,  \gamma(T)  (\frac{1}{2} + \frac{\epsilon}{\sqrt{m}})\}\cdot(1-1/e)$-approximation solution for $\textbf{P.0}$ in expectation where $\gamma(T)=1-(\frac{1}{2}+\epsilon\frac{e}{\pi})^T$.
\end{theorem}

%\begin{figure*}[hptb]
%\begin{center}
%\includegraphics[scale=0.45]{./m-normal}
%\hskip -0.08in
%\includegraphics[scale=0.45]{./k-normal}
%\hskip -0in
%\includegraphics[scale=0.5]{./n-normal}
%\caption{SMTC achieves superior expected utility under normal distributed user preferences.}
%\label{fig:smtc-normal}
%\end{center}
%\end{figure*}
%
%\begin{figure*}[hptb]
%\begin{center}
%\includegraphics[scale=0.45]{./m-uniform}
%\hskip -0.08in
%\includegraphics[scale=0.45]{./k-uniform}
%\hskip -0in
%\includegraphics[scale=0.5]{./n-uniform}
%\caption{SMTC achieves superior expected utility under uniform distributed user preferences.}
%\label{fig:smtc-uniform}
%\end{center}
%\end{figure*}

\paragraph{Discussion on Scenarios with More than Two Candidates}
We next discuss the case if we allowed to keep  $l\geq 2$ candidate solutions. In this extension, our aim is to select $l$ candidate solutions, $S_1, \cdots, S_l$, and the utility of each user-specific function is determined by the superior solution among these  candidates. Hence, our problem can be formulated as $\max_{S_1, \cdots, S_l \subseteq \Omega}\sum_{i\in[m]} \max\{f_i(S_1), \cdots, f_i(S_l)\}$ subject to  $|S_1|\leq k, \cdots, |S_l|\leq k$
where $k$ is the size constraint of a feasible solution. To tackle this challenge, we can utilize our enumeration-based partition algorithm (Algorithm \ref{alg:LPP0}) to find an approximate solution. The procedure involves enumerating all possible ways to partition the set $[m]$ into $l$ groups. For each partition, we employ a state-of-the-art $(1-1/e)$-approximation algorithm to solve the maximization problem within each group. This process generates $l$ sets, and we then choose the best $l$ sets among all partitions as the final solution. By following the same argument used to prove Theorem \ref{lem:c1}, we can show that this approach guarantees an $(1-1/e)$-approximation solution.

%\section{Conclusion}
%We note that the traditional submodular maximization approach focuses on selecting a subset of items to maximize a single submodular function. However, real-world scenarios often involve multiple user-specific functions. This necessitates a shift in our objective towards selecting sets of items that perform well across all user-specific functions. To this end, we introduce the concept of personalized submodular maximization with two candidate solutions. Our objective then becomes selecting the best pair of candidates to maximize the sum of the utilities of user-specific functions, thereby providing a (partial) personalized item selection strategy. Our study demonstrates, both theoretically and experimentally, that personalized submodular maximization with two candidates offers a balanced approach, combining the benefits of efficiency and personalization. This approach paves the way for further advancements in applications such as personalized assortment planning.
%
%In the future, we would like to develop more efficient approximation algorithms with tighter performance guarantees for large-scale problems with many user types. Additionally, it would be interesting to apply our framework to other applications such as personalized recommendation systems, targeted advertising, and customized content delivery.
\bibliographystyle{splncs04}
\bibliography{reference}

\end{document}